\documentclass[10pt,twocolumn,letterpaper]{article}

\usepackage[pagenumbers]{cvpr}
\usepackage{fontawesome5}
\pdfmapfile{+fontawesome5.map}
\usepackage{microtype}

\usepackage{float}
\usepackage{colortbl}

\definecolor{cvprblue}{rgb}{0.21,0.49,0.74}
\usepackage[breaklinks,colorlinks,allcolors=cvprblue,hyperfootnotes=false]{hyperref}

\def\confName{CVPR}
\def\confYear{2027}

\title{GlobalPuzzle: Globally Informed Genetic Search for Eroded Jigsaw Puzzle Reassembly}
\author{%
  \textbf{Lunyan Wang}\textsuperscript{1,*}\quad
  \textbf{Chuanzhi Xu}\textsuperscript{1,*\textdagger}\quad
  \textbf{Ziyuan Tao}\textsuperscript{2}\quad
  \textbf{Weidong Cai}\textsuperscript{1} \\
  \textsuperscript{1}The University of Sydney\quad
  \textsuperscript{2}Macquarie University\quad
  \href{https://github.com/Na2SO4-7258/global-puzzle}{\faGithub\ Project Page}
}

\begin{document}
\maketitle
{
\renewcommand{\thefootnote}{}
\makeatletter
\renewcommand{\@makefntext}[1]{\noindent#1}
\makeatother
\footnotetext{%
  \raggedright
  \textsuperscript{*}Equal contribution.\\
  \textsuperscript{\textdagger}Corresponding author:
  \texttt{chuanzhi.xu@sydney.edu.au}
}
}

\begin{abstract}
Jigsaw puzzle reassembly remains a highly challenging task, particularly when piece boundaries are irregularly eroded to simulate damage to cultural artifacts. On the GAP benchmark, existing methods achieve a perfect assembly accuracy of only 0.3\% on $5\times5$ puzzles, whereas humans can solve jigsaw puzzles with remarkably high accuracy. Humans typically approach puzzle solving by identifying potentially adjacent pieces, evaluating the current assembly, and iteratively performing complex operations such as adding pieces, replacing incorrectly placed pieces, and moving entire connected groups of pieces. However, existing methods face a fundamental limitation: approaches that leverage global visual information typically predict piece arrangements directly, making it difficult to explicitly perform such complex iterative operations, while search based methods can rearrange pieces but primarily rely on pairwise boundary compatibility, lacking global visual information to evaluate and guide the search process.
To bridge this gap, we propose GlobalPuzzle, a jigsaw puzzle reassembly framework that integrates global canvas level evaluation, action guidance, and genetic search. By incorporating global visual information into the evaluation and optimization of candidate assemblies, GlobalPuzzle aims to resolve ambiguous adjacency relationships that are difficult to determine from irregularly eroded piece boundaries alone, enabling flexible, globally informed reconstruction.
\end{abstract}

\section{Introduction}

Jigsaw puzzle reassembly with missing visual information is challenging due to the enormous search space and irregularly eroded piece boundaries~\cite{shahar2026gap}.
Related reconstruction problems arise in fragmented wall paintings~\cite{sizikova2018wall}.
An $N \times N$ puzzle has $(N^2)!$ possible arrangements, making the probability of randomly solving a $5 \times 5$ puzzle only $1/25! \approx 6.44695 \times 10^{-26}$.
Boundary erosion disrupts visual continuity, while smaller pieces provide less information for determining their positions and adjacency relationships~\cite{bridger2020eroded,son2016consensus,paumard2020deepzzle}.
Even advanced AI agents such as Claude Sonnet 5.5 and Codex (GPT-6 Astra) require approximately 10 minutes on average to solve a $3 \times 3$ GAP puzzle and nearly cannot reconstruct $5 \times 5$ puzzles correctly.

\begin{figure}[t]
    \centering
    \includegraphics[width=\linewidth]{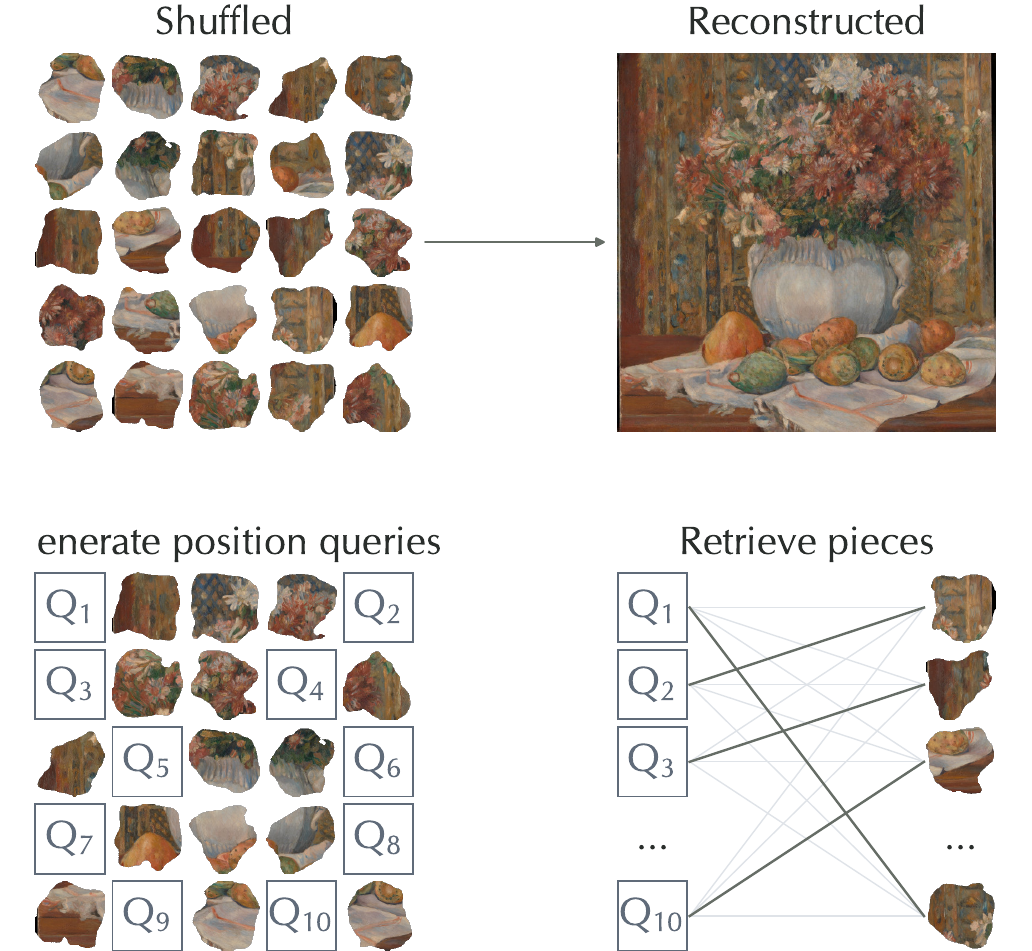}
    \caption{
      Fragment reconstruction: the top left shows randomly shuffled irregular pieces, the top right shows the image reconstructed using LaMa after correct assembly, the bottom left illustrates position query generation, and the bottom right shows query-based piece retrieval in GlobalPuzzle.
    }
    \label{fig:agent_comparison}
\end{figure}

Existing methods face two major limitations. Many deep learning approaches exchange fragment features and directly predict or refine arrangements~\cite{scarpellini2024diffassemble,liu2024jpdvt,kim2025fcvit,shahar2026gap}, but do not provide a visual fitness score for the assembled canvas or explicit operations for relocating connected groups. Exchanging fragment features does not itself evaluate whether the spatially rendered canvas forms a coherent image. Combinatorial solvers can rearrange pieces and preserve fragment relationships~\cite{pomeranz2011greedy,sholomon2013genetic}, but commonly rely on pairwise boundary compatibility, which becomes unreliable under severe erosion. Learned tree search evaluates partial assemblies~\cite{paumard2023alphazzle}, but its placement actions do not explicitly relocate connected groups.

Genetic puzzle solvers use crossover to combine structures from two candidate assemblies into a new one, preserving shared fragment connections and internal group relationships~\cite{sholomon2013genetic}. These observations suggest that preserving connected piece groups while allowing their structures to be revised could enable more flexible search. Furthermore, global canvas information offers an opportunity to guide piece placement and evaluate candidate assemblies beyond local boundary compatibility.

Motivated by these observations, we propose GlobalPuzzle, which couples explicit fragment group operations with visual guidance from assembled canvases. GlobalPuzzle has two networks. The Policy Network uses the partial canvas to guide placement, while the Evaluation Network scores complete canvases to guide genetic search. Crossover preserves and relocates shared connections, while mutation randomly changes fragment placements, allowing incorrect structures to be revised. Figure~\ref{fig:agent_comparison} illustrates the reconstruction task and position querying. Grid pooling and position to fragment matching allow the same trained model to process different canvas sizes, fragment sizes, and grid dimensions. 

Our contributions can be summarized as:
\begin{itemize}
    \item We propose GlobalPuzzle, a unified reassembly framework that integrates established connected group inheritance with learned global canvas evaluation and placement guidance.
    \item GlobalPuzzle further incorporates a reliable evaluation network that supports arbitrary canvas and fragment sizes and assesses assemblies solely through global visual semantics, without requiring pixel alignment across fragment boundaries.
    \item GlobalPuzzle accommodates different image dimensions, fragment sizes, and rectangular grids without retraining, and improves reconstruction of severely eroded puzzles over the evaluated baselines, reaching perfect assembly accuracies of 68.6\% on GAP-3 and 19.4\% on GAP-5.
\end{itemize}

\section{Related Work}

Learning-based jigsaw puzzle reassembly methods broadly fall into two categories: direct arrangement prediction methods and learning-guided search methods.

\subsection{Direct Arrangement Prediction Methods}

Spatial context prediction~\cite{doersch2015context} and jigsaw pretraining~\cite{noroozi2016jigsaw} learn visual representations from relative positions and tile arrangements. DeepPermNet~\cite{santacruz2017deeppermnet} predicts relaxed permutation matrices using Sinkhorn normalization. This differentiable relaxation enables joint learning of visual features and ordering predictions. JigsawGAN~\cite{li2022jigsawgan} couples permutation classification with image generation and adversarial supervision. FCViT~\cite{kim2025fcvit} uses a Vision Transformer~\cite{dosovitskiy2021vit} to regress fragment coordinates, avoiding a predefined permutation set. PuzLM~\cite{elkin2026puzlm} quantizes fragments into discrete tokens and predicts their reconstructed order with a sequence to sequence language model.

Other methods refine arrangements through successive predictions. DiffAssemble~\cite{scarpellini2024diffassemble} represents fragments as graph nodes and learns to denoise their translations and rotations. JPDVT~\cite{liu2024jpdvt} uses diffusion Transformers to recover positional encodings conditioned on visual content, including settings with missing pieces. PuzzleFlow~\cite{shahar2026gap} combines Vision Transformer features with discrete flow matching to update fragment positions conditioned on the current arrangement and flow time.

\begin{figure*}[!t]
    \centering
    \includegraphics[width=\textwidth]{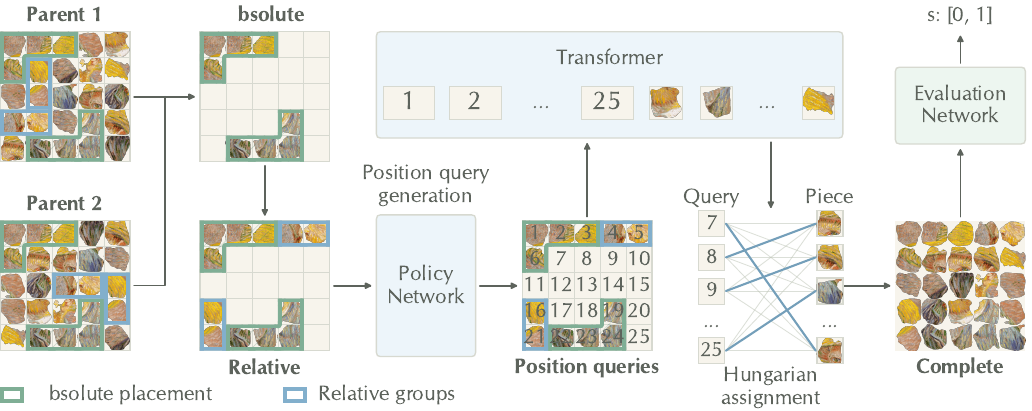}
    \caption{
      The proposed GlobalPuzzle framework. Offspring inherit common absolute placements (Absolute) and shared relative fragment relationships (Relative) from parent assemblies. The Policy Network generates position queries, which are matched with fragment features from the Piece Encoder to complete the assembly. The Evaluation Network then scores the complete canvas to guide genetic search.
    }
    \label{fig:all}
\end{figure*}

These approaches use fragment interactions or image semantic supervision to reason beyond individual boundaries. However, training an image generator or exchanging fragment features does not supply a learned fitness function that evaluates complete candidate canvases during search. Furthermore, iterative position updates do not explicitly identify connected groups, preserve their internal relationships, and relocate them as units. GlobalPuzzle combines this structural operation with global visual evaluation and placement guidance, enabling global assessment of assemblies while preserving correct piece relationships.
\subsection{Compatibility Based and Learned Search}

Compatibility based solvers separate adjacency estimation from arrangement optimization. Approaches include graphical model inference~\cite{cho2010probabilistic}, particle filtering~\cite{yang2011particle}, quadratic programming~\cite{andalo2012quadratic}, and successive linear programming~\cite{yu2016linear}. Pomeranz et al.~\cite{pomeranz2011greedy} combine greedy placement with segment rearrangement; Paikin and Tal~\cite{paikin2015missing} address mixed puzzles with missing pieces. Gallagher~\cite{gallagher2012orientation} uses gradient compatibility and constrained tree assembly for unknown orientations. Son et al. aggregate consistent cycles into larger loops~\cite{son2014loops} and grow geometric consensus to reduce dependence on unreliable local matches~\cite{son2016consensus}. Huroyan et al.~\cite{huroyan2020laplacian} recover rotations through the graph connection Laplacian before estimating piece positions.

Genetic solvers inherit shared segments through crossover~\cite{sholomon2013genetic}. Extensions address unknown orientations and piece faces~\cite{sholomon2014generalized}, and mixed puzzles with unknown dimensions~\cite{sholomon2014multiple}. These methods exploit compatibility and geometric constraints, but do not learn a visual fitness function over complete rendered canvases.

Learned matching improves local placement evidence. DNN-Buddies~\cite{sholomon2016dnnbuddies} learns adjacency from fragment pixels, while Paumard et al.~\cite{paumard2018local} predict relative positions using local feature co-occurrences. Deepzzle~\cite{paumard2020deepzzle} combines relative position prediction with shortest path optimization, including missing or extraneous fragments. Bridger et al.~\cite{bridger2020eroded} inpaint gaps between candidate neighbors and use a discriminator to classify adjacency. These approaches retain local fragment comparisons as the basis for assembly.

Alphazzle~\cite{paumard2023alphazzle} instead guides Monte Carlo tree search with policy and value networks that process the current reassembly. Its actions place the next fragment into an empty position. This provides visual guidance for construction, but does not move a connected group as a unit. GlobalPuzzle combines complete canvas evaluation with structural inheritance, connected group relocation, and placement guidance, allowing arrangements to be revised during search, thereby improving puzzle reconstruction accuracy.
\section{Methodology}

\subsection{Framework Overview}
\label{sec:overview}

 Given $M=N^2$ unordered image fragments $\mathcal{P}=\{I_i\}_{i=1}^{M}$, we seek a permutation $\pi$ assigning one fragment to each position of an $N\times N$ grid. Let $C(\pi)$ denote the resulting RGB canvas. GlobalPuzzle searches for the assembly maximizing a learned quality function:
\begin{equation}
    \hat\pi=\arg\max_{\pi\in\mathfrak{S}_M}F_\theta(\pi),\qquad
    F_\theta(\pi)=\sigma\bigl(A_\theta(C(\pi))\bigr).
    \label{eq:gp-objective}
\end{equation}
Figure~\ref{fig:all} illustrates the GlobalPuzzle framework. Given independently LaMa-inpainted fragments, genetic crossover inherits common absolute placements (Absolute) and shared relative fragment relationships (Relative) from parent assemblies. The Policy Network generates position queries from the partial canvas, while the Piece Encoder extracts fragment features. Position-to-fragment matching then completes the offspring, which the Evaluation Network scores to guide genetic search.

LaMa~\cite{suvorov2022lama} processes each eroded fragment independently without ground-truth information, providing dense RGB inputs without guaranteeing continuity between adjacent fragments.

\subsection{Policy Network}
\label{sec:placement-policy}

\begin{figure*}[!t]
    \centering
    \includegraphics[width=\textwidth]{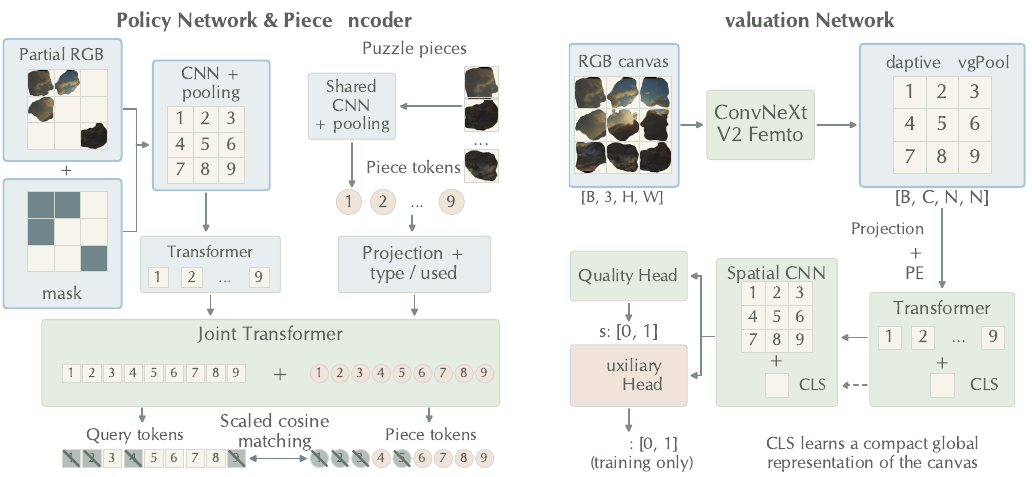}
    \caption{
       The specific structure of the Policy Network and the Evaluation Network.
    }
    \label{fig:abc}
\end{figure*}

The policy in Figure~\ref{fig:abc} encodes all fragments into tokens $c_i$ and extracts a query $q_p$ for each grid position from the partial RGB canvas and its occupancy map. A joint Transformer updates both token sets. The resulting pairwise scores are:
\begin{equation}
S_{p,i}=\tau\,
\frac{\langle W_q\tilde q_p,W_c\tilde c_i\rangle}
{\|W_q\tilde q_p\|_2\,\|W_c\tilde c_i\|_2},
\label{eq:gp-policy}
\end{equation}
where $\tau$ is learned. One forward pass produces an $M\times M$ position to fragment score matrix.

Training uses correct and perturbed partial canvases. When all occupied fragments are correctly placed, each empty position receives a one-hot ground-truth target. Otherwise, occupied neighbors vote for fragments that preserve their ground-truth directional relationships. Normalized vote counts define targets $t_{p,i}$, and the policy minimizes:
\begin{equation}
\mathcal L_B=-\frac1{|\mathcal Q|}\sum_{p\in\mathcal Q}\sum_{i=1}^{M}
 t_{p,i}\log\operatorname{softmax}(S_{p,:})_i,
\label{eq:gp-b-loss}
\end{equation}
where $\mathcal Q$ contains supervised empty positions. During search, occupied positions and used fragments are masked. Fragment embeddings are cached across candidates.

\subsection{Genetic Search with Global Fitness}
\label{sec:ga}
\par\noindent{\textbf{Initialization and Selection.}}
The policy generates an initial population of complete permutations. Each is evaluated by $F_\theta$. The strongest individuals are retained as elites, while parents are sampled with fitness weighted probabilities:
\begin{equation}
P(\pi_k)=\frac{(F_\theta(\pi_k)+\epsilon)^\beta}
{\sum_j(F_\theta(\pi_j)+\epsilon)^\beta}.
\label{eq:gp-selection}
\end{equation}
\par\noindent{\textbf{Structural Crossover.}}
Following genetic solvers~\cite{sholomon2013genetic}, shared oriented adjacencies define connected groups $G$. Given parents $\pi^a,\pi^b$, only groups of at least two fragments with identical absolute placements for all members are copied, each with probability $0.5$:
\begin{equation}
s_p=\pi^a(p),\quad\forall p\text{ such that }\pi^a(p)\in G.
\label{eq:gp-absolute}
\end{equation}
Other groups preserve relative fragment relationships. When a fragment $i$ is assigned a position, its connected partners $j$ can be placed by translation:
\begin{equation}
r_s(j)=r_s(i)+r_G(j)-r_G(i),\quad j\in G,
\label{eq:gp-relative}
\end{equation}
provided that all placements remain within the grid and do not conflict with occupied cells.

\par\noindent{\textbf{Policy Guided Completion.}}
For each distinct partial offspring, the policy generates the score matrix $S$ once. The Hungarian algorithm~\cite{kuhn1955hungarian} solves a one-to-one assignment between vacant positions $\mathcal U$ and unused fragments $\mathcal R$:
\begin{equation}
\mathcal M^*=\arg\max_{\mathcal M\in\mathfrak M(\mathcal U,\mathcal R)}
\sum_{(p,i)\in\mathcal M}S_{p,i}.
\label{eq:gp-hungarian}
\end{equation}
Here $\mathfrak M(\mathcal U,\mathcal R)$ denotes all one-to-one complete matchings. Matched pairs are committed in descending score order. After each commitment, compatible parent shared groups are inherited. Any remaining slots are matched using the same policy scores, yielding a complete permutation.

\par\noindent{\textbf{Mutation and Refinement.}}
Nonelite offspring undergo swaps, three fragment cycles, or block exchanges. Periodic local search evaluates all two position swaps of the best candidate and accepts the highest scoring improvement. The final output is the best assembly encountered during genetic search.

\subsection{Evaluation Network}
\label{sec:global-eval}

\par\noindent{\textbf{Architecture.}}
The evaluator in Figure~\ref{fig:abc} processes the complete canvas using a multiscale CNN~\cite{woo2023convnextv2} followed by a Transformer~\cite{vaswani2017attention}. Features pooled to the puzzle grid retain spatial organization, while a global token and spatial readout provide complementary representations. The network outputs a quality logit $z_\theta=A_\theta(C(\pi))$ and a fitness score $\hat Q(\pi)=\sigma(z_\theta)$. This score is denoted by $s\in[0,1]$ in Figures~\ref{fig:all} and~\ref{fig:abc}, with higher values indicating better predicted assembly quality. An auxiliary head estimates absolute accuracy during training.
\par\noindent{\textbf{Quality Supervision.}}
Let $\pi^*$ be the ground truth, $r_\pi(i)$ the grid coordinate of fragment $i$, and $\mathcal E^*$ the set of $2N(N-1)$ oriented ground-truth adjacencies $(i,j,\delta)$. We define:
\begin{align}
\mathrm{AA}(\pi)&=\frac1M\sum_{p=1}^{M}\mathbf1[\pi(p)=\pi^*(p)],\label{eq:gp-aa}\\
\mathrm{SRA}(\pi)&=\frac1{|\mathcal E^*|}\sum_{(i,j,\delta)\in\mathcal E^*}
\mathbf1[r_\pi(j)-r_\pi(i)=\delta],\label{eq:gp-sra}\\
Q(\pi)&=0.6\,\mathrm{AA}(\pi)+0.4\,\mathrm{SRA}(\pi).
\label{eq:gp-quality}
\end{align}
We generate training assemblies with controlled displacement counts and structural transformations, including block exchanges, row and column shifts, and mixed mutations.
\par\noindent{\textbf{Training Objective.}}
The evaluator is trained with logit regression, score calibration, within puzzle ranking, hard negative mining, and auxiliary AA regression:
\begin{equation}
\mathcal L_A=\mathcal L_{\rm reg}+0.5\mathcal L_{\rm cal}
+0.75\mathcal L_{\rm rank}+0.5\mathcal L_{\rm hard}
+0.2\mathcal L_{\rm AA}.
\label{eq:gp-a-loss}
\end{equation}
Regression targets are $\operatorname{logit}(\operatorname{clip}(Q,0.01,0.99))$. For candidates $i,j$ of the same puzzle with $Q_i>Q_j$, the ranking term penalizes:
\begin{equation}
\ell_{ij}=\operatorname{softplus}\bigl(3(Q_i-Q_j)-z_i+z_j\bigr),
\label{eq:gp-rank}
\end{equation}
with larger weights for greater quality differences and structural negatives. The hard negative term penalizes the most overestimated incorrect candidate relative to the correct assembly.

\subsection{Final Reconstruction}
After determining the final piece arrangement, we reconstruct the canvas using the original, non-inpainted fragments. LaMa is then applied to the assembled canvas to fill the missing regions using global visual context, producing the final reconstructed image shown in Figure~\ref{fig:agent_comparison}. This step uses a pretrained model without additional training and does not participate in the puzzle-solving process.

%-------------------------------------------------------------------------

\section{Experiments and Results}
\label{sec:experiments}

\subsection{Experimental Setup}
\label{sec:exp-setup}

\noindent\textbf{Datasets and Evaluation Metrics.}
We evaluate GlobalPuzzle on GAP-3 ($3\times3$) and GAP-5 ($5\times5$)~\cite{shahar2026gap}, following the official splits of 14,000 training, 3,000 validation, and 3,000 test puzzles for each dataset. Each fragment is independently inpainted using LaMa to provide RGB inputs for the Evaluation Network. The Piece Encoder processes individual fragments, while the Policy Network uses a partial canvas and occupancy mask.

We report perfect assembly accuracy (PA) and absolute accuracy (AA), measuring the percentage of fully reconstructed puzzles and correctly placed fragments, respectively. All experiments use an NVIDIA RTX 4060 GPU, with average reconstruction times of 9 and 44 seconds per puzzle on GAP-3 and GAP-5, respectively.

\noindent
\textbf{Network Training.}
GlobalPuzzle contains three learned components: the \emph{Evaluation Network}, the \emph{Policy Network}, and the \emph{Piece Encoder}. The Piece Encoder extracts reusable fragment representations. The Policy Network combines these representations with a partial canvas to predict fragment scores for candidate positions, while the Evaluation Network predicts the quality of a complete assembled canvas.

The Evaluation Network is supervised using the quality target $Q=0.6\,\mathrm{AA}+0.4\,\mathrm{SRA}$ described in Sec.~\ref{sec:global-eval}. Training candidates include correct assemblies, permutations with controlled numbers of misplaced fragments, and structural perturbations. Its objective combines score regression, calibration, candidate ranking, hard negative supervision, and auxiliary AA prediction. We train the Evaluation Network for 50 epochs using AdamW~\cite{loshchilov2019adamw}, a batch size of 16, a learning rate of $3\times10^{-4}$, weight decay of $0.05$, and a three epoch warmup. Mixed precision training~\cite{micikevicius2018mixed} is enabled.

The Policy Network is trained on partially completed canvases containing both correct placements and deliberately perturbed placements. Supervision follows the target construction in Sec.~\ref{sec:placement-policy}: empty positions receive the corresponding ground-truth fragment target when the context is correct, or a distribution derived from neighboring fragment relationships when the context is perturbed. The prediction loss is computed against all fragments, while invalid assignments are excluded when completing a puzzle. Fragment representations produced by the Piece Encoder are cached and reused during genetic search.

Detailed network architectures and training procedures are provided in Appendices~\ref{sec:supp-method} and~\ref{sec:supp-training}, respectively.

\begin{table}[t]
\centering
\caption{Comparison on GAP-3 ($3\times3$) and GAP-5 ($5\times5$). PA and AA are reported as percentages.}
\label{tab:main_results}
\begingroup
\renewcommand{\arraystretch}{1.15}
\setlength{\tabcolsep}{5pt}
\small
\begin{tabular}{@{}lcccc@{}}
\toprule
& \multicolumn{2}{c}{GAP-3} & \multicolumn{2}{c}{GAP-5} \\
\cmidrule(lr){2-3}\cmidrule(lr){4-5}
Method & PA $\uparrow$ & AA $\uparrow$ & PA $\uparrow$ & AA $\uparrow$ \\
\midrule
\rowcolor{black!10}
\multicolumn{5}{@{}l@{}}{\textit{Classical}} \\
Greedy~\cite{pomeranz2011greedy} (CVPR'11) & 0.0 & 11.6 & 0.0 & 4.1 \\
GA~\cite{sholomon2013genetic} (CVPR'13) & 0.0 & 11.1 & 0.0 & 11.1 \\
\addlinespace[2pt]
\rowcolor{black!10}
\multicolumn{5}{@{}l@{}}{\textit{Deep Learning}} \\
JigsawGAN~\cite{li2022jigsawgan} (TIP'22) & 4.6 & 45.3 & 0.0 & 18.0 \\
DiffAssemble~\cite{scarpellini2024diffassemble} (CVPR'24) & 16.4 & 50.5 & 0.0 & 21.9 \\
JPDVT~\cite{liu2024jpdvt} (CVPR'24) & 0.0 & 11.2 & 0.0 & 3.9 \\
Elkin et al.~\cite{elkin2026puzlm} (ECCV'26) & 0.0 & 14.8 & 0.0 & 7.8 \\
FCViT~\cite{kim2025fcvit} (ESWA'25) & 25.2 & 60.7 & 0.0 & 20.4 \\
PuzzleFlow~\cite{shahar2026gap} (CVPR'26) & 28.5 & 62.9 & 0.3 & 29.1 \\
\midrule
GlobalPuzzle (Ours) & \textbf{68.6} & \textbf{81.6} & \textbf{19.4} & \textbf{48.7} \\
\bottomrule
\end{tabular}
\endgroup
\end{table}

\begin{figure*}[!t]
    \centering
    \includegraphics[width=\textwidth]{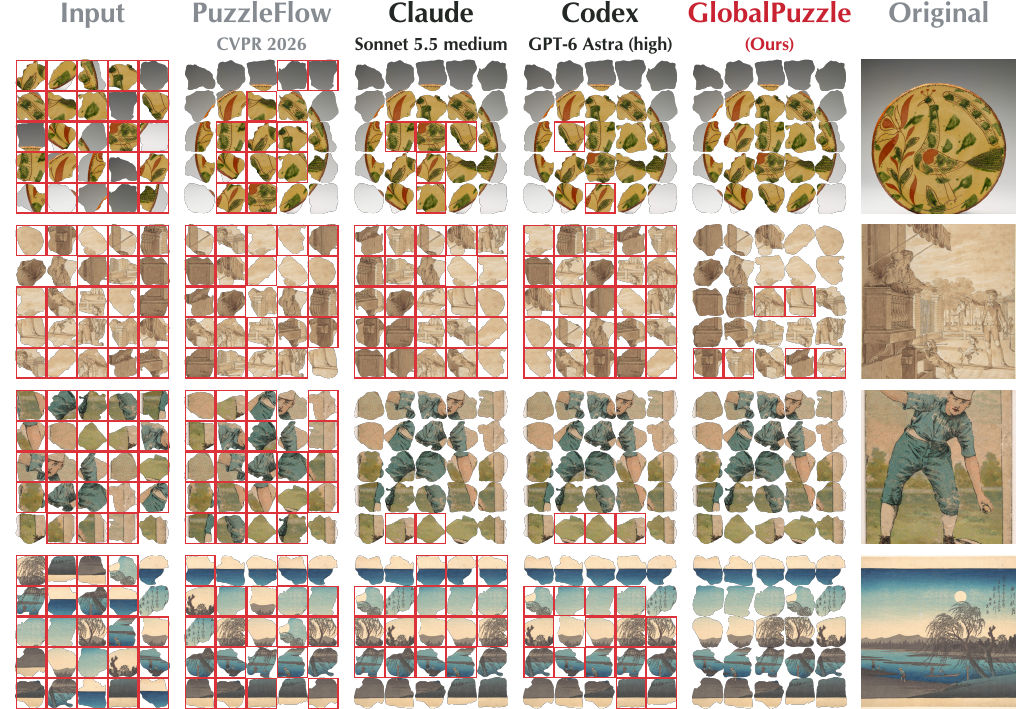}
    \caption{
       Visual comparison of the restoration for each method. Codex and Claude performed puzzle reconstruction through their respective web interfaces, while PuzzleFlow~\cite{shahar2026gap} was evaluated using its officially released pretrained weights.
    }
    \label{fig:compare}
\end{figure*}

\noindent
\textbf{Genetic Search Configuration.}
Unless otherwise specified, we use a population of 256 individuals, 30 generations, and four elites per generation. Parents are selected using fitness weighted sampling with exponent $\beta=4$. Genetic crossover preserves shared absolute placements and shared relative fragment relationships, followed by completion using the Policy Network and assignment optimization. We use random completion probability $0.05$ per vacant position and mutation probability $0.20$ per nonelite offspring. Local refinement is performed every five generations, with at most four accepted improvement steps per invocation. To maintain consistency across comparisons, the same search budget is used for the corresponding ablation variants. Appendix~\ref{sec:supp-search} provides implementation and inference-efficiency details.

\noindent
\textbf{Comparison Methods.}
We compare GlobalPuzzle with classical optimization methods and learning based approaches. The classical group includes greedy assembly~\cite{pomeranz2011greedy} and a conventional genetic algorithm~\cite{sholomon2013genetic} based on local fragment compatibility. The learning based group includes JigsawGAN~\cite{li2022jigsawgan}, DiffAssemble~\cite{scarpellini2024diffassemble}, JPDVT~\cite{liu2024jpdvt}, FCViT~\cite{kim2025fcvit}, PuzzleFlow~\cite{shahar2026gap}, and the method of Elkin et al.~\cite{elkin2026puzlm}. These baselines cover direct arrangement prediction, learned reconstruction, and compatibility driven optimization. We additionally examine representative attempts by general purpose multimodal agents to illustrate the challenges of explicit fragment manipulation and iterative reconstruction. Since agent attempts are analyzed as qualitative case studies, they are not treated as directly comparable benchmark measurements.

\subsection{Comparison with Existing Methods and General Purpose Agents}
\label{sec:exp-comparison}

Table~\ref{tab:main_results} compares reconstruction accuracy on GAP-3 and GAP-5.GlobalPuzzle achieves mean PA and AA of $68.6\%$ and $81.6\%$ on GAP-3, respectively. On the more challenging GAP-5 benchmark, the corresponding means are $19.4\%$ and $48.7\%$. Per-seed results and standard deviations are reported in Appendix~\ref{sec:supp-seeds}. These results show improvements in complete reconstruction and fragment placement relative to the baselines listed in the table. The decline from GAP-3 to GAP-5 also indicates that larger assemblies remain substantially more difficult, despite the use of global image information and iterative search.

Figure~\ref{fig:compare} compares reconstruction results from PuzzleFlow~\cite{shahar2026gap}, general purpose multimodal agents Codex and Claude, and our GlobalPuzzle. PuzzleFlow is evaluated using its officially released pretrained weights, while Codex and Claude perform reconstruction through their respective web interfaces. While existing methods and agents can recover partial image structures, they often struggle to produce fully correct arrangements. In contrast, GlobalPuzzle combines global visual evaluation with iterative genetic search to refine fragment placements and improve reconstruction accuracy. These examples reveal misplaced fragments in individual puzzles, complementing the aggregate metrics without quantifying agent performance. Appendix~\ref{sec:supp-visual} presents additional successful reconstructions and failure cases.

\begin{table}[t]
\centering
\caption{Ablation study on GAP-3 and GAP-5. PA and AA are reported as percentages.}
\label{tab:ablation}
\begingroup
\renewcommand{\arraystretch}{1.15}
\resizebox{\columnwidth}{!}{
\begin{tabular}{@{}l@{\hspace{14pt}}cccc@{}}
\toprule
& \multicolumn{2}{c}{GAP-3 (\%)}
& \multicolumn{2}{c}{GAP-5 (\%)} \\
\cmidrule(lr){2-3}\cmidrule(lr){4-5}
Method & PA $\uparrow$ & AA $\uparrow$ & PA $\uparrow$ & AA $\uparrow$ \\
\midrule
Random Evaluation Network & 1.9 & 25.2 & 0.0 & 9.6 \\
Random Policy Network & 65.1 & 73.4 & 0.0 & 16.9 \\
No Crossover & 60.0 & 75.4 & 10.4 & 42.5 \\
No Swapping & 63.5 & 76.2 & 11.3 & 42.7 \\
No Crossover and Swapping & 54.5 & 70.7 & 0.1 & 9.3 \\
\midrule
GlobalPuzzle (Full) & \textbf{68.6} & \textbf{81.6} & \textbf{19.4} & \textbf{48.7} \\
\bottomrule
\end{tabular}
}
\endgroup
\end{table}

\subsection{Ablation Studies}
\label{sec:exp-ablation}

Table~\ref{tab:ablation} isolates the learned networks and genetic operations under the same datasets, metrics, and search budget. We replace each network's outputs with random scores or disable crossover, swapping, or both, while retaining the other components.

\noindent
\textbf{Evaluation Network and Policy Network.}
The Evaluation Network directs elite retention, parent selection, local refinement, and the final output. Random evaluation reduces GAP-3 PA/AA from $68.6/81.6\%$ to $1.9/25.2\%$ and GAP-5 from $19.4/48.7\%$ to $0.0/9.6\%$. Even with the learned policy and genetic operators intact, uninformative fitness cannot reliably retain good candidates. Candidate diversity therefore cannot compensate for losing the signal that steers evolution.

\noindent\textbf{Policy guidance becomes critical as the grid grows.}
Random policy scores retain legal assignments but remove learned placement guidance. On GAP-3, PA decreases from $68.6\%$ to $65.1\%$ and AA from $81.6\%$ to $73.4\%$. Inherited placements and connected groups constrain the $9!$ arrangements, allowing population diversity to partly compensate for weak guidance without guaranteeing exhaustive search. On GAP-5, random policy scores reduce PA to $0.0\%$ and AA to $16.9\%$. With $25!$ unconstrained arrangements, the same budget cannot supply enough useful candidates through unguided completion. The policy therefore becomes essential for exploring the larger search space.

\noindent
\textbf{Genetic Search Operations.}
Crossover and swapping supply candidates for evaluation. Crossover recombines inherited placements and connected groups, while swapping revises fragment positions within an assembly. Removing crossover leaves swapping available and yields GAP-3/GAP-5 PA of $60.0/10.4\%$; removing swapping leaves crossover available and yields $63.5/11.3\%$. Each remaining operator can still propose changes that the evaluator selects, so the framework retains partial effectiveness despite lower accuracy.

Removing both operators eliminates these complementary routes for revising candidates. GAP-5 PA/AA then collapse to $0.1/9.3\%$, compared with $19.4/48.7\%$ for the full model. GAP-3 retains $54.5/70.7\%$, consistent with the smaller search space being easier to cover. Together, these results show that evaluation chooses which candidates survive, while policy guidance and genetic variation determine which candidates become available. Reliable evaluation needs a continuing supply of useful alternatives, particularly on GAP-5.

\subsection{Analysis of the Evaluation Network and Policy Network}
\label{sec:exp-network-analysis}

\begin{figure}[!t]
    \centering
    \includegraphics[width=\linewidth]{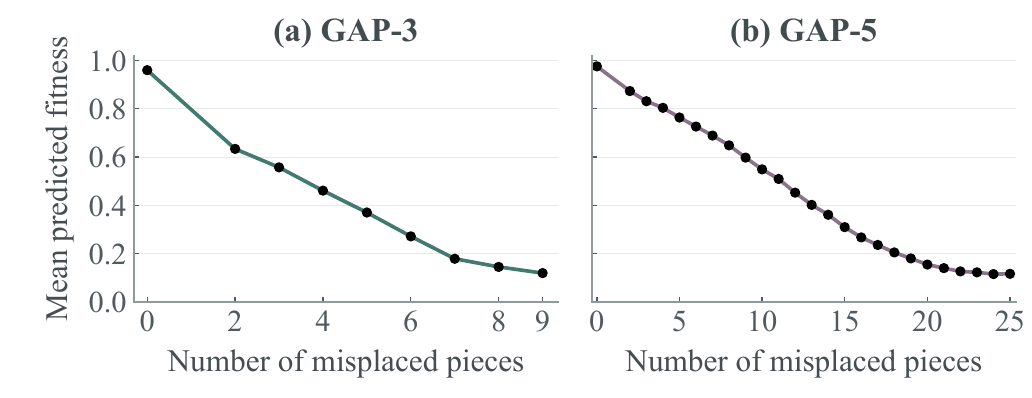}
    \caption{
       Evaluation Network: Mean predicted fitness versus the number of misplaced pieces on GAP-3 and GAP-5.
    }
    \label{fig:A}
\end{figure}

\begin{figure}[!t]
    \centering
    \includegraphics[width=\linewidth]{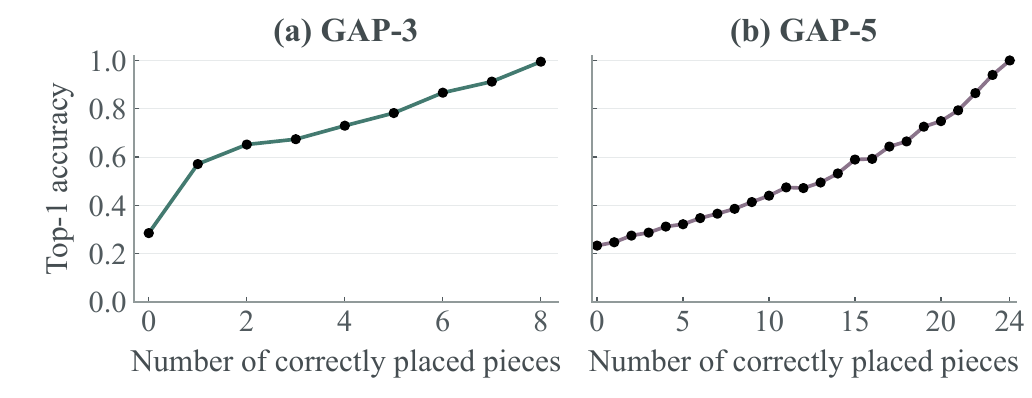}
    \caption{
       Policy Network: Top-1 accuracy versus the number of correctly placed pieces on GAP-3 and GAP-5.
    }
    \label{fig:BC}
\end{figure}

\noindent
\textbf{Evaluation Network.}
Figure~\ref{fig:A} plots mean predicted fitness against the number of misplaced fragments on GAP-3 and GAP-5. This diagnostic examines whether scores respond to intermediate reconstruction quality, beyond distinguishing complete solutions from incorrect assemblies. Such sensitivity matters because selection and refinement must compare partially correct candidates throughout search. The curves summarize average behavior; they do not establish that every candidate pair is ranked correctly. The random evaluation ablation complements this diagnostic by measuring the downstream effect of removing informative fitness from the complete solver.

\begin{figure}[!t]
    \centering
    \includegraphics[width=\linewidth]{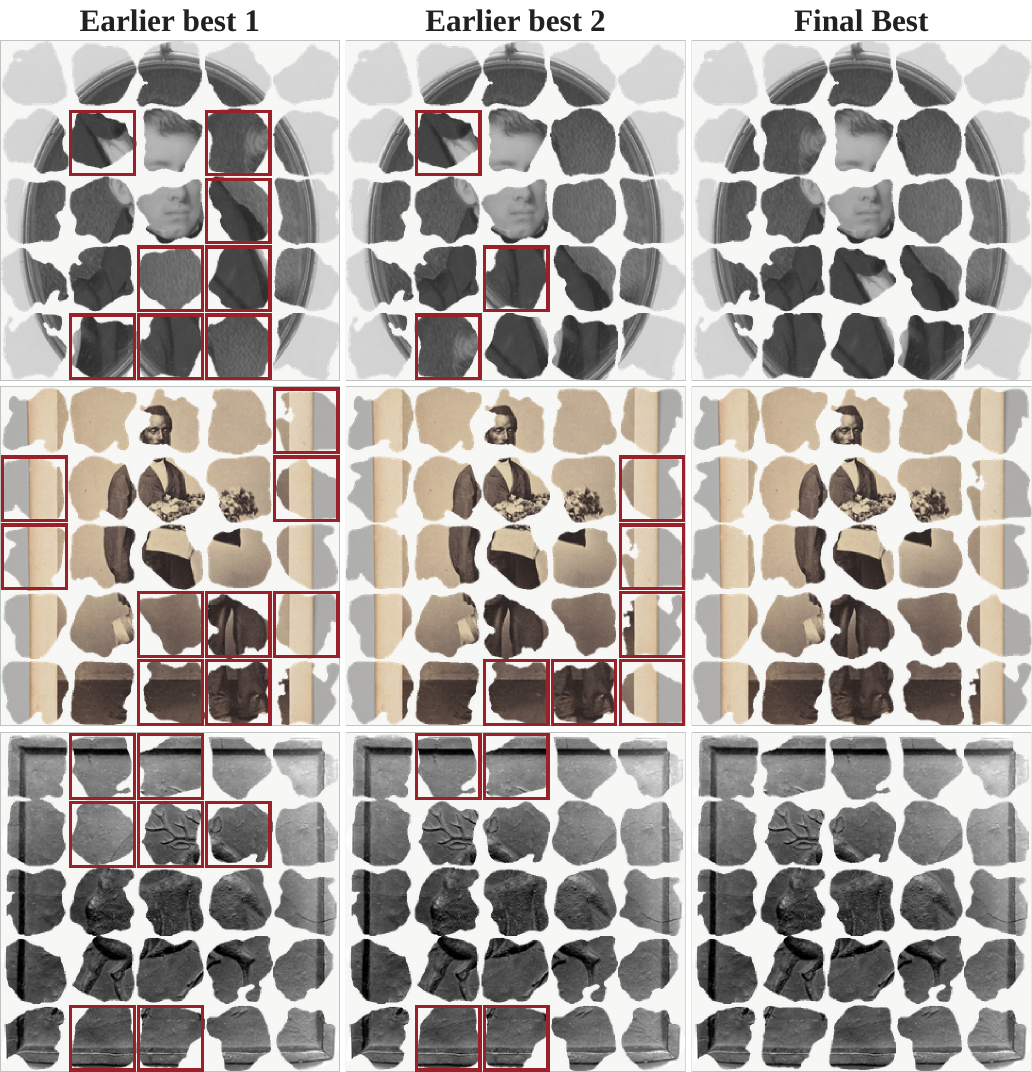}
    \caption{
       Genetic search progression from earlier candidates to the final reconstruction. Red outlines mark misplaced fragments in each candidate assembly.
    }
    \label{fig:ga}
\end{figure}

\noindent
\textbf{Policy Network.}
Figure~\ref{fig:BC} reports the Policy Network's top one fragment prediction accuracy over supervised empty positions. A prediction is correct when the highest scoring fragment matches the target, with accuracy computed from the prediction scores before the assignment stage. This isolates the learned placement signal from the effects of the Hungarian algorithm and later genetic search.

Results are reported separately for GAP-3 and GAP-5 and grouped by the number of correctly placed fragments to examine the effect of available visual context. The random policy ablation measures a different property: whether population search can compensate for losing this placement signal. Strong recovery on GAP-3 therefore does not make the policy redundant; its contribution becomes decisive when the larger GAP-5 search space limits candidate coverage.

\subsection{Analysis of the Genetic Search Process}
\label{sec:exp-search-analysis}

We analyze the evolution of candidate assemblies throughout genetic search. Figure~\ref{fig:ga} presents the best candidate from selected generations together with its reconstruction quality. This visualization illustrates how search can improve a partially correct arrangement through changes in individual fragment positions and the translation of connected fragment groups. Such group operations are particularly useful when several fragments have a consistent internal arrangement but occupy an incorrect region of the canvas.

\subsection{Real World Photo Reconstruction and Generalization}
\label{sec:exp-generalization}

\begin{figure}[!t]
    \centering
    \includegraphics[width=\linewidth]{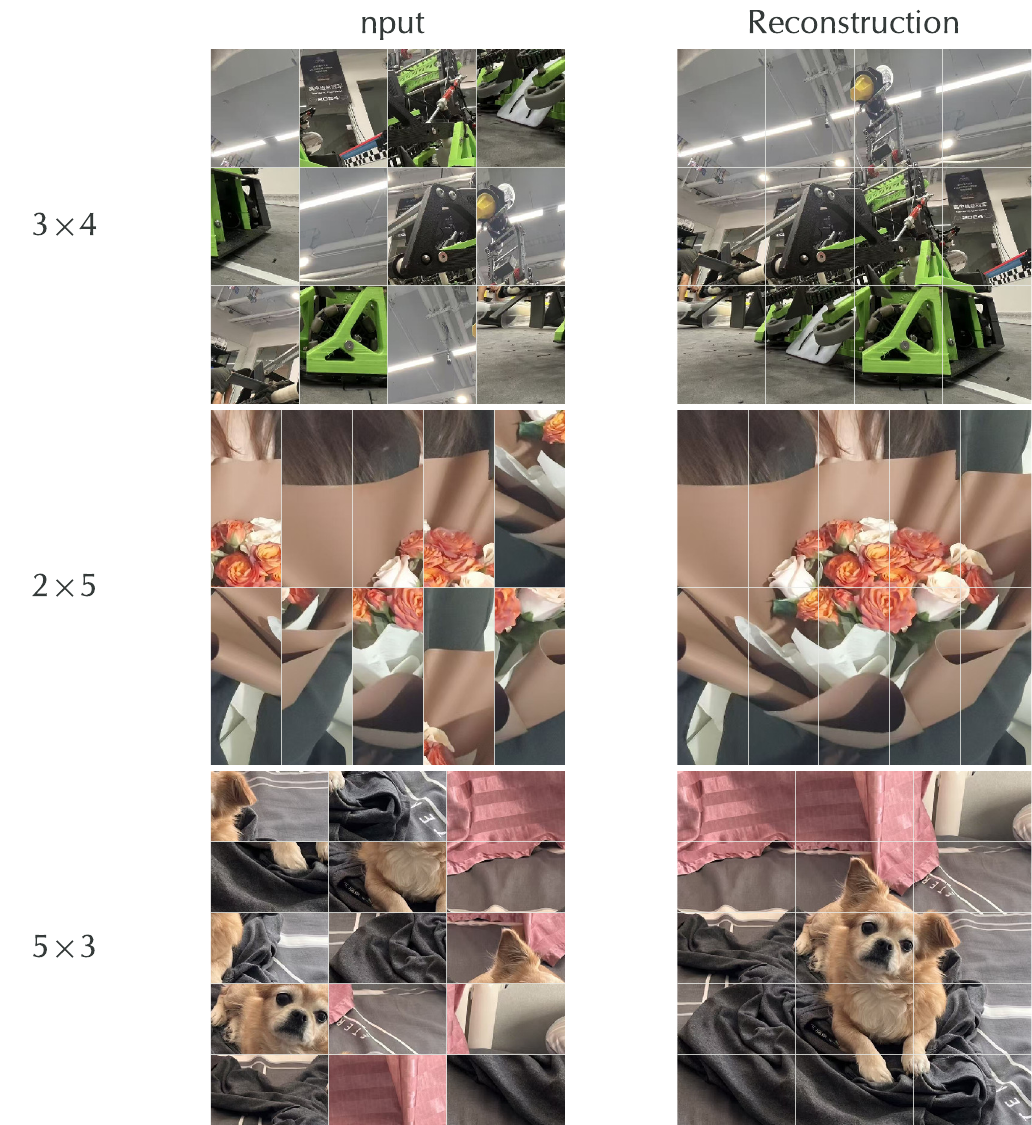}
    \caption{
       Reconstructing images from different numbers and sizes of fragments in daily life.
    }
    \label{fig:live}
\end{figure}

To evaluate generalization beyond the GAP benchmarks, we apply GlobalPuzzle to real world photographs and puzzle configurations of different sizes, including rectangular grids such as $3\times4$, $2\times5$, and $5\times3$. The pretrained networks are applied without additional training or fine tuning. Figure~\ref{fig:live} presents representative reconstruction results across different scenes and puzzle configurations.

These experiments examine the ability of GlobalPuzzle to generalize to unseen image content and grid dimensions. Position queries follow the target grid, while the Evaluation Network scores the resulting complete canvas. This allows the same learned components to support reconstruction beyond the square configurations used during training.

\section{Conclusion}
GlobalPuzzle integrates global canvas evaluation, placement guidance, and genetic search for eroded puzzles. Global semantics guide fragment replacement and connected group relocation without requiring pixel alignment. Experiments show improved reconstruction on GAP-3 and GAP-5 and generalization across canvas sizes and grid dimensions without retraining. Ablations show that reliable evaluation is essential for selecting useful assemblies, while placement guidance and genetic variation supply promising candidates, especially as the search space grows. Together, these components preserve useful fragment relationships while allowing incorrect structures to be revised, linking local placement decisions to the coherence of the complete assembly. The lower accuracy on GAP-5 shows that larger assemblies remain difficult despite global visual guidance. Future work will explore faster search for arbitrary fragment shapes and layouts while retaining the ability to revise connected groups under limited computational budgets.

% References follow the main text without a forced submission page break.
{\small
\bibliographystyle{ieeenat_fullname}
\bibliography{references}
}

\clearpage
\appendix
\setcounter{table}{0}
\setcounter{figure}{0}
\setcounter{equation}{0}
\renewcommand{\thesection}{\Alph{section}}
\renewcommand{\thetable}{S\arabic{table}}
\renewcommand{\thefigure}{S\arabic{figure}}
\renewcommand{\theequation}{S\arabic{equation}}

\twocolumn[{
\begin{center}
{\LARGE\bfseries Supplementary Material}
\end{center}
\vspace{1em}
}]

\section{Robustness to Random Seeds}
\label{sec:supp-seeds}
We evaluate the robustness of GlobalPuzzle to stochastic genetic search using three random seeds (42, 43, and 2026) on GAP-3 and GAP-5. The pretrained Evaluation Network and Policy Network, evaluation samples, and all search hyperparameters are kept identical across runs. Only the random seed controlling population initialization, parent selection, connected group placement, and mutation is varied. Each experiment uses a population of 256 individuals and a maximum of 30 generations.

Table~\ref{tab:seed} reports perfect assembly accuracy (PA) and absolute accuracy (AA) for each seed, together with their mean and standard deviation. The full-model results in the main text and Tables~\ref{tab:main_results} and~\ref{tab:ablation} use these means. These experiments assess the sensitivity of reconstruction performance to different stochastic search trajectories while isolating the influence of random seeds from network training and other experimental settings.

\begin{table}[t]
\centering
\caption{Robustness to random seeds on GAP-3 and GAP-5. Results are reported as percentages.}
\label{tab:seed}
\renewcommand{\arraystretch}{1.15}
\setlength{\tabcolsep}{1pt}
\hspace{-3mm}
\begin{tabular*}{\columnwidth}{@{\extracolsep{\fill}}lcccc@{}}
\toprule
& \multicolumn{2}{c}{GAP-3} & \multicolumn{2}{c}{GAP-5} \\
\cmidrule(lr){2-3}\cmidrule(lr){4-5}
Seed & PA & AA & PA & AA \\
\midrule
42 & 69.2 & 80.2 & 19.5 & 48.4 \\
43 & 68.5 & 82.8 & 19.1 & 49.6 \\
2026 & 68.0 & 81.8 & 19.7 & 48.0 \\
\midrule
M$\pm$S & $68.6\pm0.6$ & $81.6\pm1.3$ & $19.4\pm0.3$ & $48.7\pm0.8$ \\
\bottomrule
\end{tabular*}
\end{table}

\section{Method Details}
\label{sec:supp-method}
\subsection{Evaluation Network}
\label{sec:supp-evaluation}
\noindent{\textbf{Backbone and Spatial Tokens.}} The Evaluation Network scores a completed RGB canvas using ConvNeXt V2 Femto pretrained on ImageNet-1K. Features from three backbone stages are extracted after resizing the canvas to $64N\times64N$, where $N$ is the puzzle grid width. ImageNet normalization is applied before the backbone. Each feature map is adaptively average-pooled to $N\times N$, and the pooled features are concatenated and projected to 192 channels with a linear layer and LayerNorm. This yields one token per puzzle position.

% GlobalPuzzle supplementary tables: one dedicated, full-width float page.
% Required package in the main preamble: \usepackage{booktabs}
% Insert with \input{globalpuzzle_supp_tables_no_clearpage}
% Each table retains its own caption, number, and reference label.
% Floating [p] permits preceding text to fill the page naturally.
\begin{table*}[p]
\centering
\renewcommand{\arraystretch}{1.18}
\setlength{\tabcolsep}{8pt}

\begin{minipage}{\textwidth}
\centering
\caption{Evaluation Network architecture. $N$ denotes the puzzle grid width.}
\label{tab:supp-evaluator-arch}
\begin{tabular}{@{}lll@{}}
\toprule
Module & Configuration & Output \\
\midrule
Backbone & ConvNeXt V2 Femto, feature stages 1--3 & Multiscale feature maps \\
Grid pooling & Adaptive average pooling to $N\times N$ & $N\times N$ per scale \\
Projection & Linear, LayerNorm, positional embedding & $N^2\times192$ \\
Transformer & 3 layers, 6 heads, FFN dimension 384 & $(N^2+1)\times192$ \\
Spatial CNN & Conv 192--96, residual block, Conv 96--48 & $48\times5\times5$ \\
Fusion & Flattened spatial features and CLS token & 1392 \\
Quality Head & 1392--256--1 & Fitness logit \\
Auxiliary AA Head & 1392--128--1 & AA logit \\
\bottomrule
\end{tabular}
\end{minipage}

\vspace{0.048\textheight}

\begin{minipage}{\textwidth}
\centering
\caption{Policy Network and Piece Encoder architectures. Query and piece tokens interact through the joint Transformer.}
\label{tab:supp-policy-arch}
\begin{tabular}{@{}lll@{}}
\toprule
Module & Configuration & Output \\
\midrule
Canvas encoder & RGB and occupancy input; shallow 64, deep 256 channels & Feature maps \\
Grid pooling & Both scales to $2N\times2N$, grouped in $2\times2$ cells & 1280 per position \\
Query projection & LayerNorm, Linear, GELU & $N^2\times256$ \\
Query Transformer & 3 layers, 8 heads, FFN dimension 1024 & $N^2\times256$ \\
Piece Encoder & Shared CNN, $4\times4$ pooling, MLP & $N^2\times128$ \\
Piece projection & Linear 128--256 & $N^2\times256$ \\
Joint Transformer & 2 layers, 8 heads, FFN dimension 1024 & $2N^2\times256$ \\
Matching heads & Residual, LayerNorm, Linear & $N^2\times128$ each \\
Pairwise scoring & Learned-scale cosine similarity & $N^2\times N^2$ \\
\bottomrule
\end{tabular}
\end{minipage}

\vspace{0.050\textheight}

\begin{minipage}[t]{0.49\textwidth}
\centering
\setlength{\tabcolsep}{3pt}
\caption{Training and search hyperparameters used for GlobalPuzzle.}
\label{tab:supp-config}
\begin{tabular}{@{}lr@{}}
\toprule
Setting & Value \\
\midrule
Evaluation / Policy epochs & 50 / 50 \\
Evaluation / Policy batch size & 16 / 8 \\
Learning rate / weight decay & $3\times10^{-4}$ / 0.05 \\
Warmup epochs & 3 \\
Backbone learning-rate multiplier & 0.1 \\
Population / maximum generations & 256 / 30 \\
Elites / roulette exponent & 4 / 4.0 \\
Absolute group inheritance probability & 0.50 \\
Random placement per empty slot & 0.05 \\
Mutation per nonelite child & 0.20 \\
Local search interval / accepted steps & 5 / 4 \\
Policy / Evaluation inference batch & 32 / 32 \\
Policy / Evaluation cache capacity & 8192 / 32768 \\
\bottomrule
\end{tabular}
\end{minipage}\hfill
\begin{minipage}[t]{0.49\textwidth}
\centering
\setlength{\tabcolsep}{6pt}
\caption{Structural negative sampling and full-offspring mutation distributions.}
\label{tab:supp-sampling}
\begin{tabular}{@{}lr@{}}
\toprule
Operation & Probability \\
\midrule
\multicolumn{2}{@{}l}{\emph{Evaluation training negatives}} \\
Block exchange & 0.40 \\
Row or column shift & 0.30 \\
Row or column exchange & 0.20 \\
Mixed mutation & 0.10 \\
\midrule
\multicolumn{2}{@{}l}{\emph{Complete offspring mutation}} \\
Two-piece swap & 0.60 \\
Three-piece cycle & 0.25 \\
Two-cell block exchange & 0.15 \\
\bottomrule
\end{tabular}
\end{minipage}
\end{table*}

Learnable position embeddings are stored on a $5\times5$ reference grid and interpolated bicubically for other grid dimensions. A learned CLS token and its positional embedding are prepended to the sequence. A three-layer Transformer encoder uses six attention heads, a feedforward dimension of 384, GELU activations, pre-normalization, and dropout of 0.1. It exchanges information across all puzzle positions and produces both spatial tokens and a global CLS representation.

\noindent{\textbf{Spatial Readout and Fitness.}} The $N^2$ spatial tokens are reshaped to an $N\times N$ feature map. A $3\times3$ convolution reduces the channels from 192 to 96, followed by a depthwise-pointwise residual block and a $1\times1$ convolution reducing the channels to 48. GroupNorm and SiLU are used in the spatial branch. Its output is resized bilinearly to $5\times5$ and flattened to 1200 dimensions. Concatenating this vector with the 192-dimensional CLS token produces a 1392-dimensional feature. The Quality Head applies a 1392--256--1 MLP with SiLU and dropout of 0.1. A separate 1392--128--1 MLP predicts auxiliary absolute accuracy (AA) during training. The sigmoid of the quality logit is used as the genetic-search fitness.

\subsection{Policy Network}
\label{sec:supp-policy}
The Policy Network receives a four-channel partial canvas consisting of RGB values and a binary occupancy mask. A convolutional stem and shallow branch produce 64-channel local features, while a deeper residual CNN produces 256-channel features. Both branches are pooled to $2N\times2N$ and concatenated. Grouping each $2\times2$ neighborhood yields a 1280-dimensional representation per puzzle position. LayerNorm, a linear projection, and GELU map these features into 256-dimensional position queries.

Two-dimensional sinusoidal positional encodings are computed from normalized row and column coordinates for the current grid. A learned occupied-or-empty state embedding is added to each query. A three-layer query Transformer with eight attention heads, a feedforward dimension of 1024, and dropout of 0.1 updates the $N^2$ position queries using the partial canvas context.

\subsection{Piece Encoder and Joint Matching}
\label{sec:supp-piece}
The Piece Encoder applies a shared CNN independently to each LaMa-inpainted RGB fragment. Four convolutional stages expand the feature channels from 32 to 64, 128, and 256, using stride-two downsampling and residual blocks. The feature map is adaptively pooled to $4\times4$. An MLP maps the flattened 4096-dimensional vector through 512 dimensions to a 128-dimensional piece embedding.

Piece embeddings are projected into the 256-dimensional query space. Learned token-type embeddings identify queries and pieces, while used-state embeddings mark pieces already placed. A two-layer joint Transformer with eight attention heads and a feedforward dimension of 1024 processes all $2N^2$ tokens using bidirectional self-attention. The joint outputs are added to their corresponding input embeddings and projected through separate LayerNorm and linear heads to 128-dimensional query and piece representations. The representations are $\ell_2$-normalized and matched by scaled cosine similarity. The learned logit scale is initialized to 12 and bounded to $[1,40]$.

All query and piece tokens participate in joint attention. Occupied-position and used-piece constraints are imposed when selecting placements, rather than masking the joint Transformer. The Piece Encoder outputs can be reused across candidate canvases of the same puzzle. Query encoding and joint attention are recomputed because the partial canvas and used-piece states change.

\section{Training and Data Details}
\label{sec:supp-training}
\subsection{Evaluation Network Supervision}
\label{sec:supp-eval-training}
The continuous quality target combines absolute accuracy and oriented structural reassembly accuracy as $Q=0.6\,AA+0.4\,SRA$. For each puzzle, the base candidate pool contains the correct permutation and candidates with exactly $k$ misplaced pieces for every $k\in\{2,\ldots,N^2\}$. Each incorrect candidate is formed by selecting $k$ positions and sampling a derangement on those positions. Thus, the candidate has exactly $k$ misplaced pieces, and a single misplaced piece is never sampled because it cannot occur in a complete permutation.

Four additional structural negatives are generated using block exchanges, row or column shifts, row or column exchanges, and mixed mutations. Their sampling probabilities are reported in Table~\ref{tab:supp-sampling}. These transformations retain some local structure while changing the global arrangement. Each training candidate receives its continuous $Q$ target and an AA target.

The training loss combines SmoothL1 regression on $\operatorname{logit}(\operatorname{clip}(Q,0.01,0.99))$, mean squared calibration error on the sigmoid fitness, pairwise ranking within each puzzle, hard-negative supervision, and auxiliary AA regression:
\begin{equation}
\mathcal L_A=\mathcal L_{\mathrm{reg}}+0.5\mathcal L_{\mathrm{cal}}+0.75\mathcal L_{\mathrm{rank}}+0.5\mathcal L_{\mathrm{hard}}+0.2\mathcal L_{\mathrm{AA}}.
\label{eq:supp-eval-loss}
\end{equation}
Candidate pairs are ranked according to their ground-truth quality difference using a margin scale of 3.0. Pairs involving structural negatives receive an additional weighting factor of 1.5. The hard-negative term penalizes the largest quality-adjusted score violation among incorrect candidates relative to the correct assembly.

\subsection{Policy Network Supervision}
\label{sec:supp-policy-training}
The Policy Network and Piece Encoder are trained jointly. Each training example consists of the full set of $N^2$ fragments, a partially assembled RGB canvas, its binary occupancy mask, the identities of already used fragments, and a target distribution for each supervised empty position. A partial state assigns a distinct fragment to each occupied position and leaves all other positions empty. The ground-truth permutation is used to construct the state and supervision, but is not given to either network as input.

\noindent{\textbf{Number of Placed Fragments.}} For each training example, the number $k$ of occupied positions is sampled independently. An empty canvas ($k=0$) is selected with probability 0.10, a one-fragment canvas ($k=1$) with probability 0.30, and $k$ is sampled uniformly from $\{2,\ldots,N^2-1\}$ with total probability 0.60. Empty canvases use a ground-truth one-hot target at every position. One-fragment canvases reproduce random initialization: the seed is correctly positioned with probability $1/N^2$, and otherwise a fragment is placed at an incorrect position.

\noindent{\textbf{Partial-State Construction.}} For $k\geq2$, three state types are sampled with probabilities 0.45, 0.30, and 0.25, respectively. In \emph{Random Correct} states, $k$ distinct positions are selected uniformly, and each receives its correct ground-truth fragment. In \emph{Group Correct} states, occupied positions are constructed by growing spatially connected groups from separated seeds, with two or three groups when the grid and occupancy count permit. Every selected position still receives its correct fragment. For $k=1$ and $k=N^2-1$, the grouped-position sampler falls back to randomly selected positions when separate groups cannot be formed.

In \emph{Corrupted} states, a correct partial arrangement is first constructed from either uniformly selected positions or grouped positions, each with probability 0.5. A perturbation then exchanges two occupied fragments, replaces an occupied fragment with an unused fragment, or translates a small connected subset of occupied fragments when a legal shift exists. The perturbation preserves unique fragment identities and the number of occupied positions. A corrupted sample is accepted only when at least one fragment occupies an incorrect absolute position and an unused ground-truth fragment remains eligible for a vacant position. Invalid states are resampled. These three probabilities apply to the $k\geq2$ branch; the one-fragment branch uses the separate random-initialization distribution described above.

\noindent{\textbf{Position Targets and Loss.}} If all occupied fragments are correctly placed, each empty position is supervised by the one-hot identity of its ground-truth fragment. Otherwise, occupied four-neighbors of each empty position propose fragments according to the corresponding relative offsets in the ground-truth permutation. The votes from valid neighboring anchors are normalized to form a soft target. Empty positions without a valid target do not contribute to the loss, and occupied positions are never directly supervised. For every supervised empty position, cross-entropy is computed over the complete set of $N^2$ fragment logits, including fragments already placed on the canvas. Occupancy and used-fragment masks are applied during inference assignment rather than to the training softmax.

\subsection{Optimization and Preprocessing}
\label{sec:supp-optimization}
Both networks are trained for 50 epochs with AdamW, an initial learning rate of $3\times10^{-4}$, weight decay of 0.05, and a three-epoch warmup. The Evaluation Network uses a batch size of 16 canvases, and the Policy Network uses a batch size of 8. The pretrained evaluator backbone uses 0.1 times the base learning rate. Mixed precision is enabled. Detailed configurations are listed in Table~\ref{tab:supp-config}.

Each eroded fragment is independently inpainted using LaMa before network processing, without access to the correct complete arrangement. This preprocessing supplies RGB inputs while retaining the need to infer correct adjacency and global placement. After the permutation is determined, the original fragments are reassembled and LaMa is applied to the complete canvas for final image restoration. The latter stage does not influence the predicted permutation.

\section{Genetic Search Implementation}
\label{sec:supp-search}
\subsection{Structural Crossover}
\label{sec:supp-crossover}
Given two parent permutations, directed rightward and downward adjacency relations shared by both parents define connected groups containing at least two fragments. A group qualifies for absolute inheritance only if every member occupies the same absolute position in both parents. The entire group is retained with probability 0.5. Single fragments are not inherited solely because their positions agree.

Groups not fixed by absolute inheritance are considered for relative inheritance in descending order of group size. Each group is translated without rotation to a uniformly sampled feasible location that preserves its internal offsets and does not overlap previously placed pieces. When no legal translation exists, its fragments remain available for subsequent policy completion.

\subsection{Completion, Mutation, and Refinement}
\label{sec:supp-completion}
The Policy Network scores each distinct partially assembled offspring once. Before Hungarian assignment, each vacant position receives a random unused fragment with probability 0.05. The remaining positions and fragments are then matched one-to-one. The resulting complete offspring are evaluated by the Evaluation Network. Four elites are retained, and parents are sampled with fitness-weighted roulette selection using exponent $\beta=4$.

Each nonelite complete offspring is independently mutated with probability 0.20 using a two-piece swap, a three-piece cycle, or a two-cell block exchange. The respective operation probabilities are given in Table~\ref{tab:supp-sampling}. Local refinement evaluates all two-position swaps of the best assembly and accepts the highest-fitness improvement. It runs at generation zero, every five generations, and at the final generation, with at most four accepted improvements each time. The search uses 256 individuals and up to 30 generations, and returns the best-scoring assembly encountered.

\subsection{Inference Efficiency}
\label{sec:supp-efficiency}
The Piece Encoder outputs are cached separately from the canvas-dependent joint attention results. Policy scores and Evaluation Network scores are cached per puzzle, with capacities of 8192 and 32768 entries. Inference uses batches of up to 32 candidate canvases for each network. On an NVIDIA RTX 4060, average reconstruction takes approximately 9 seconds per GAP-3 puzzle and 44 seconds per GAP-5 puzzle.

\section{Additional Visual Results}
\label{sec:supp-visual}

\subsection{Additional Method Comparisons}
\begin{figure*}[!t]
    \centering
    \includegraphics[
        width=\textwidth,
        height=0.86\textheight,
        keepaspectratio
    ]{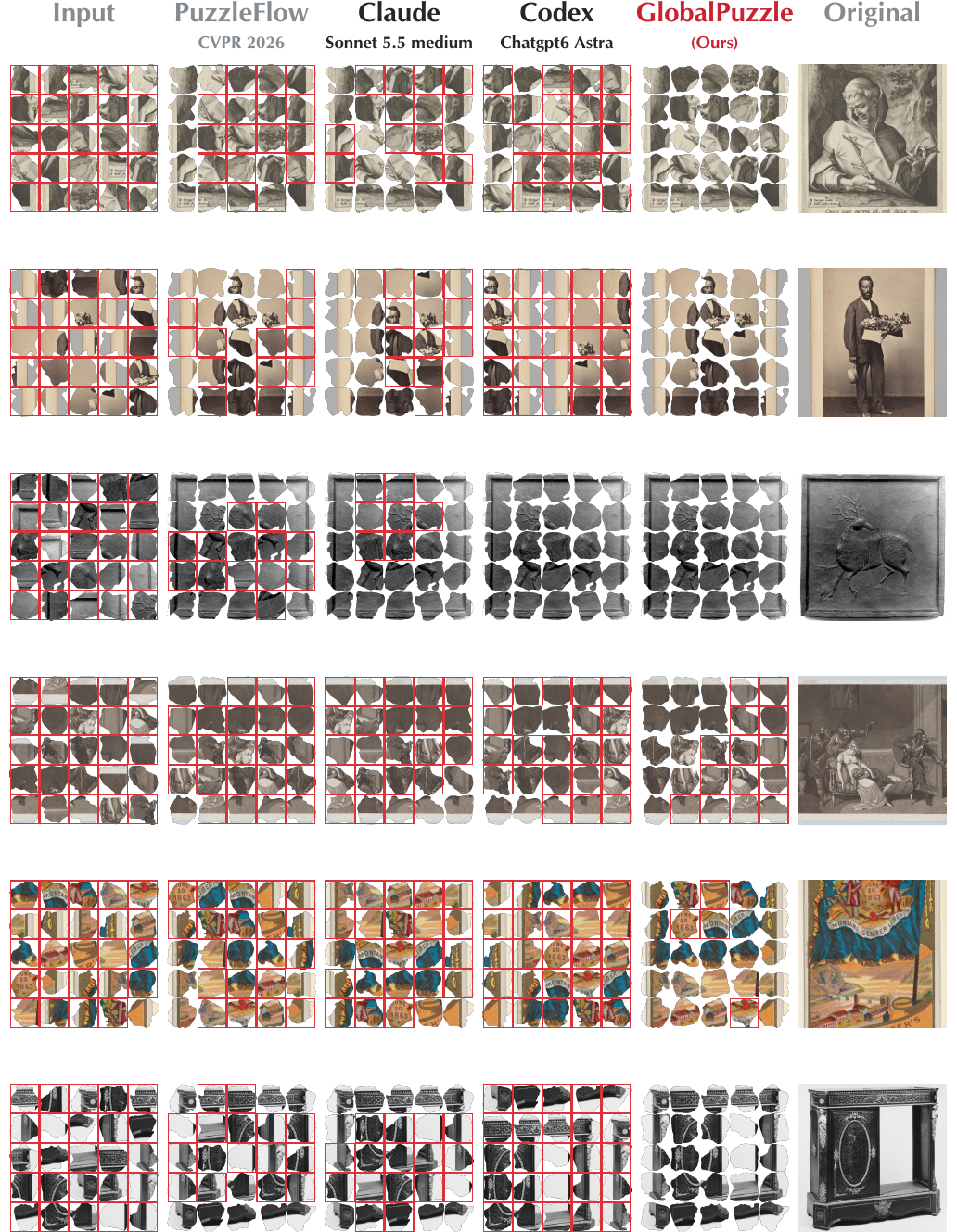}
    \caption{
        Additional compare reconstruction
    }
    \label{fig:compare1}
\end{figure*}

Figure~\ref{fig:compare1} presents additional visual comparisons between GlobalPuzzle and existing methods on GAP-3 and GAP-5. These results further demonstrate the effectiveness of GlobalPuzzle across diverse visual content and puzzle sizes.

\subsection{Additional Real-World Reconstruction Results}
\begin{figure*}[!t]
    \centering
    \includegraphics[
        width=\textwidth,
        height=0.86\textheight,
        keepaspectratio
    ]{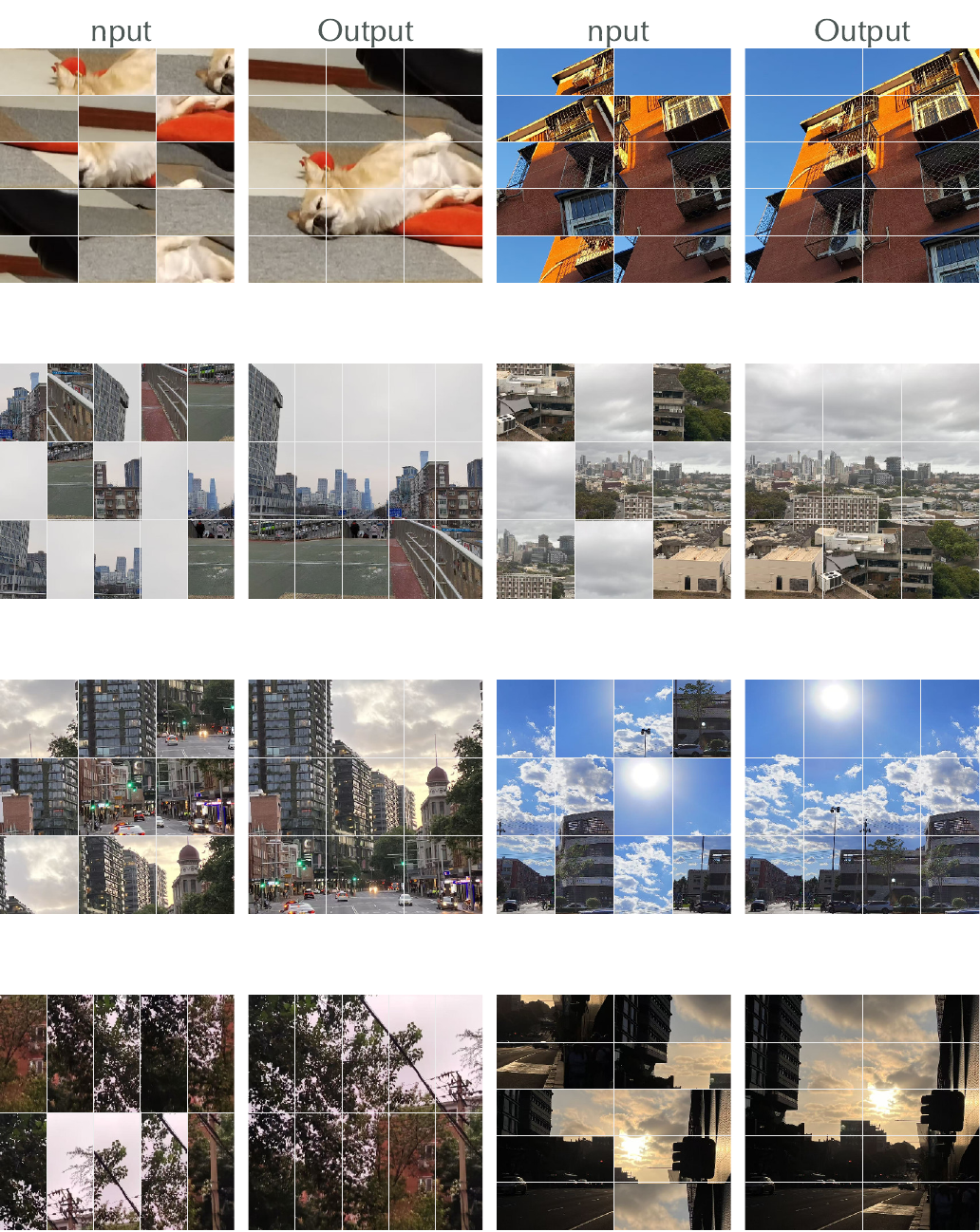}
    \caption{
        Additional examples of reconstruction live photos
    }
    \label{fig:live1}
\end{figure*}

Figure~\ref{fig:live1} presents additional reconstruction examples on real-world photographs with different grid configurations. These results demonstrate the ability of GlobalPuzzle to handle diverse image content and puzzle layouts without retraining.

\begin{figure*}[!t]
    \centering
    \includegraphics[
        width=\textwidth,
        height=0.86\textheight,
        keepaspectratio
    ]{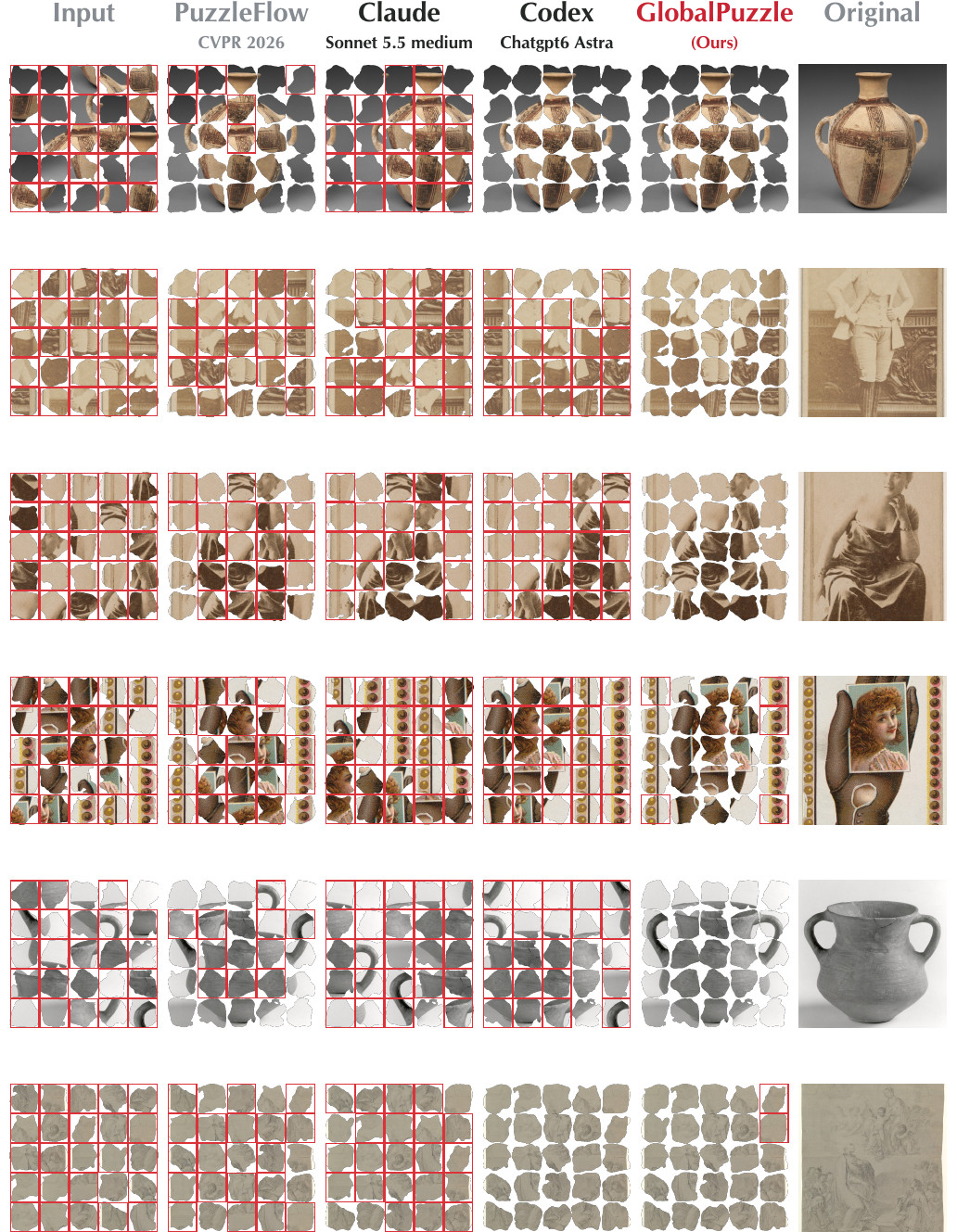}
    \caption{
        Additional compare reconstruction
    }
    \label{fig:compare2}
\end{figure*}

\begin{figure*}[!t]
    \centering
    \includegraphics[
        width=\textwidth,
        height=0.86\textheight,
        keepaspectratio
    ]{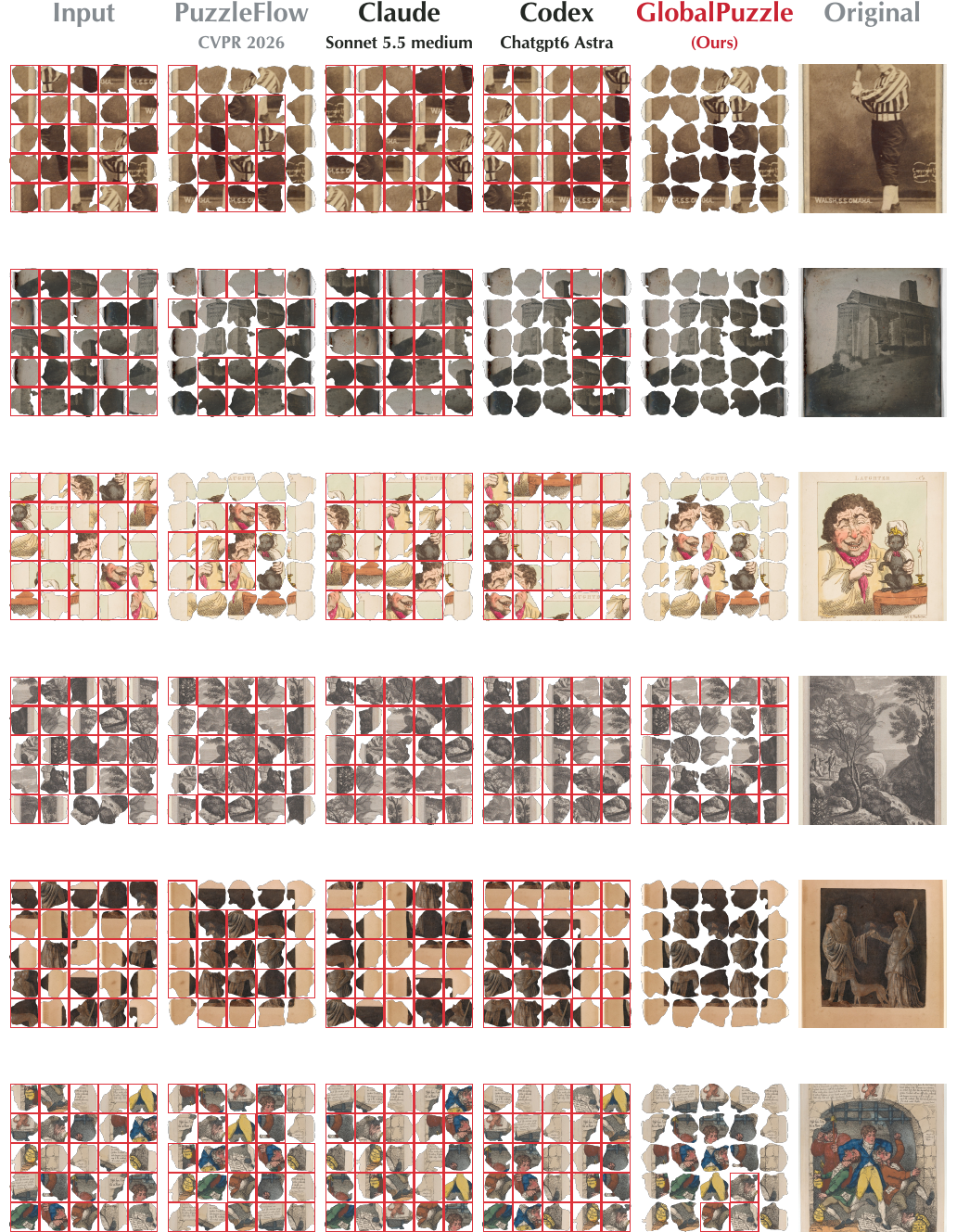}
    \caption{
        Additional compare reconstruction
    }
    \label{fig:compare3}
\end{figure*}

\begin{figure*}[!t]
    \centering
    \includegraphics[
        width=\textwidth,
        height=0.86\textheight,
        keepaspectratio
    ]{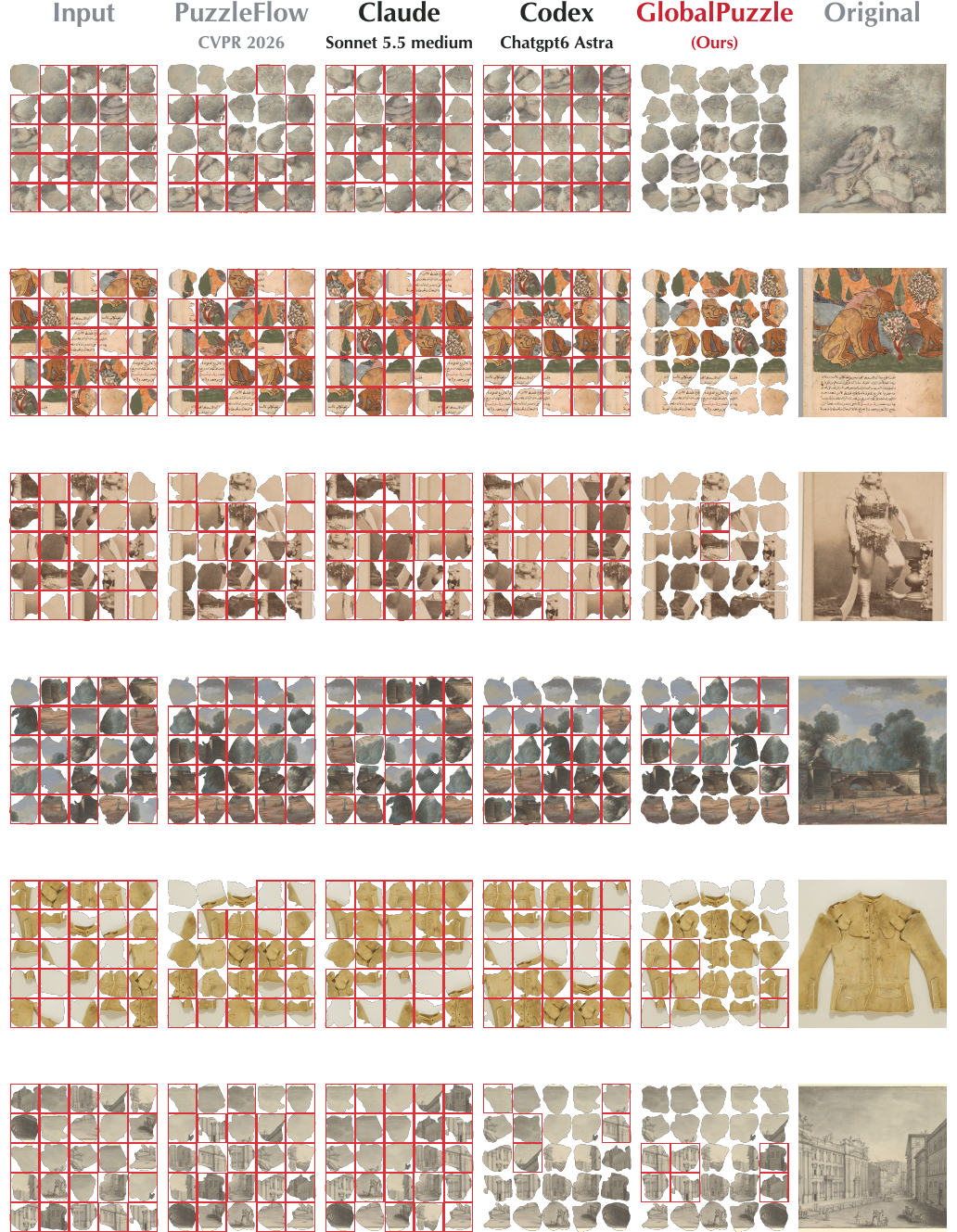}
    \caption{
        Additional compare reconstruction
    }
    \label{fig:compare4}
\end{figure*}

\begin{figure*}[!t]
    \centering
    \includegraphics[
        width=\textwidth,
        height=0.86\textheight,
        keepaspectratio
    ]{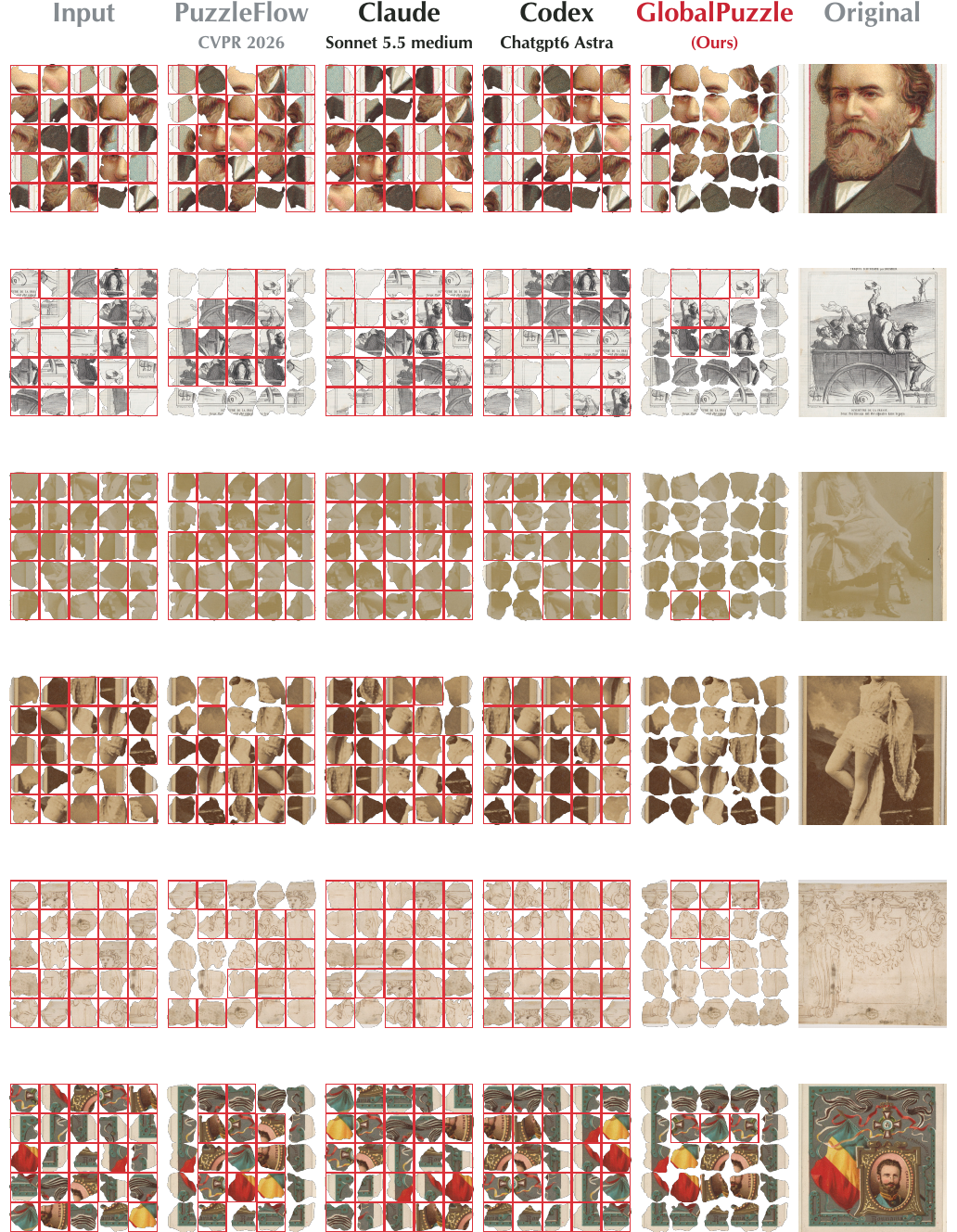}
    \caption{
        Additional compare reconstruction
    }
    \label{fig:compare5}
\end{figure*}

\begin{figure*}[!t]
    \centering
    \includegraphics[
        width=\textwidth,
        height=0.86\textheight,
        keepaspectratio
    ]{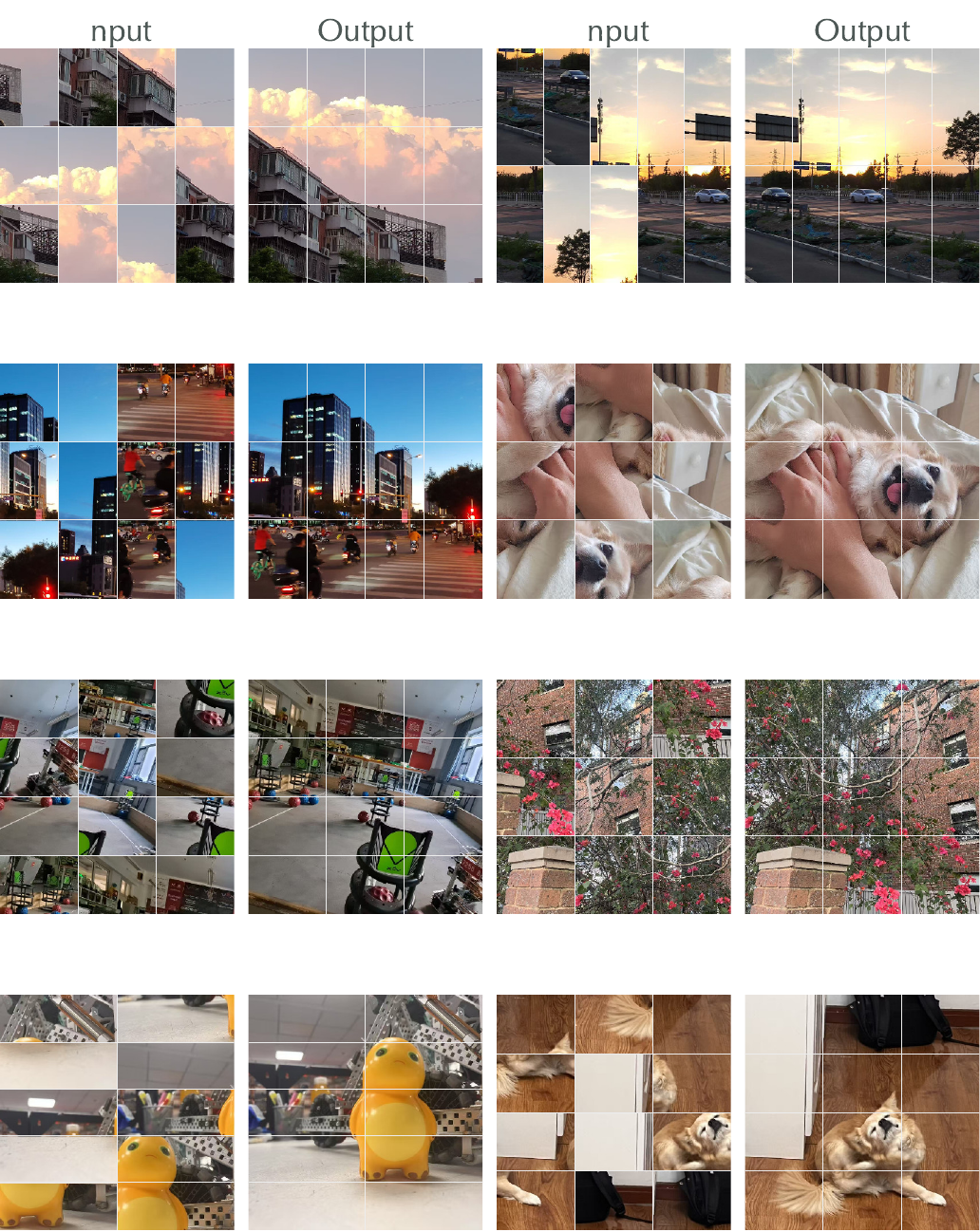}
    \caption{
        Additional examples of reconstruction live photos
    }
    \label{fig:live2}
\end{figure*}

\begin{figure*}[!t]
    \centering
    \includegraphics[
        width=\textwidth,
        height=0.86\textheight,
        keepaspectratio
    ]{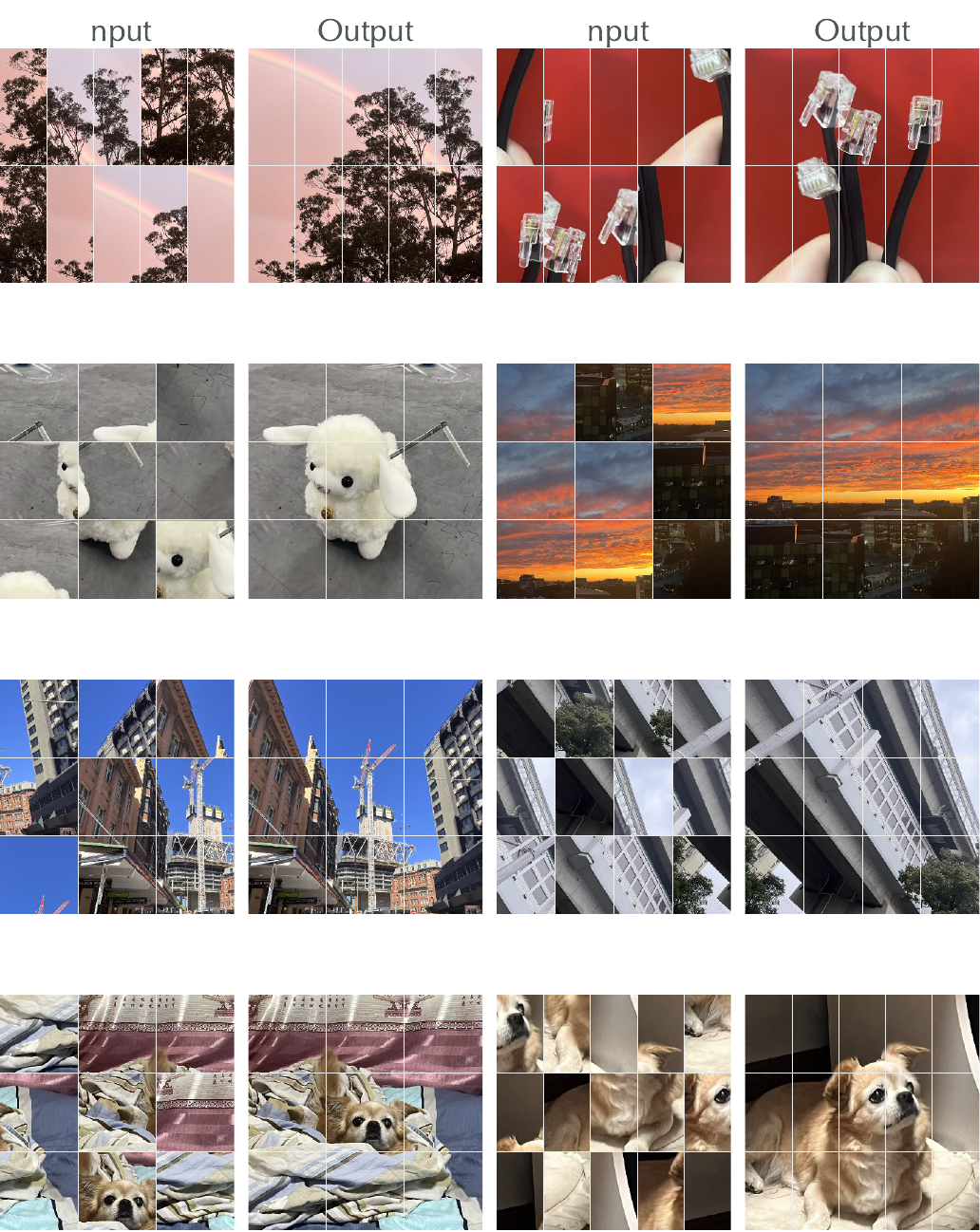}
    \caption{
        Additional examples of reconstruction live photos
    }
    \label{fig:live3}
\end{figure*}

\end{document}